\documentclass[runningheads]{llncs}
\usepackage[shortlabels]{enumitem}
\usepackage[T1]{fontenc}
\usepackage{graphicx}
\usepackage{tabularx}
\PassOptionsToPackage{hyphens}{url}
\usepackage{hyperref}
\usepackage{multirow}
\usepackage{float}
\usepackage{fancyvrb}
\graphicspath{{./images/}}

\usepackage{xspace}
\makeatletter
\DeclareRobustCommand\onedot{\futurelet\@let@token\@onedot}
\def\@onedot{\ifx\@let@token.\else.\null\fi\xspace}

\def\etal{\textit{et al}\onedot}
\makeatother

\usepackage[dvipsnames]{xcolor}

\newcommand{\RQBOX}[1]{\noindent\fbox{\parbox{0.97\linewidth}{\bfseries #1}}}

\begin{document}

\title{A deep Natural Language Inference predictor without language-specific training data}
\titlerunning{A deep NLI predictor without language-specific training data}
%
\author{Lorenzo Corradi\inst{1} \and
Alessandro Manenti\inst{1} \and
Francesca Del Bonifro\inst{1} \and
Francesco Setti\inst{2}\orcidID{0000-0002-0015-5534} \and
Dario Del Sorbo\inst{1} \orcidID{0000-0003-0390-7540}}

\authorrunning{Lorenzo Corradi \etal}
\institute{Data Science Team, Lutech S.p.A., Cinisello Balsamo, Italy
\and
Department of Engineering for Innovation Medicine, University of Verona, Verona, Italy\\
\vspace{.3em}\email{l.corradi@lutech.it}}

\maketitle              

\begin{abstract}
In this paper we present a technique of NLP to tackle the problem of inference relation (NLI) between pairs of sentences in a target language of choice without a language-specific training dataset.
We exploit a generic translation dataset, manually translated, along with two instances of the same pre-trained model --- the first to generate sentence embeddings for the source language, and the second fine-tuned over the target language to mimic the first. This technique is known as Knowledge Distillation.
The model has been evaluated over machine translated Stanford NLI test dataset, machine translated Multi-Genre NLI test dataset, and manually translated RTE3-ITA test dataset.
We also test the proposed architecture over different tasks to empirically demonstrate the generality of the NLI task. The model has been evaluated over the native Italian ABSITA dataset, on the tasks of Sentiment Analysis, Aspect-Based Sentiment Analysis, and Topic Recognition. 
We emphasise the generality and exploitability of the Knowledge Distillation technique that outperforms other methodologies based on machine translation, even though the former was not directly trained on the data it was tested over.
\end{abstract}
\section{Introduction}
\label{Introduction}
Natural Language Processing (NLP) has gained huge improvements and importance in the last years. It has many different applications as it helps in many ways human language productions understanding and analysis in an automated manner.
Natural Language Inference (NLI) is one of these applications: it is the task of determining the inference relation between two short texts written in natural language, usually defined as \emph{premise} and \emph{hypothesis}~\cite{SNLI,MNLI}. 
This implies the extraction of the meaning of the two texts and then evaluating if the \emph{Premise} (P) entails the \emph{Hypothesis} (H) (\textit{entailment} situation), if the \emph{premise} and the \emph{hypothesis} are in contradiction between each other (\textit{contradiction} situation), or if none of these two situations happen and there is no inference relation among the two texts (\textit{neutral} situation). 
This is a challenging task that requires understanding the nuances of language and context, as well as the ability to reason and make logical implications.
The relevance of this task can be easily understood by highlighting some of its possible applications. 
Common tasks based on NLI are Aspect-Based Sentiment Analysis (ABSA), Sentiment Analysis (SA), and Topic Recognition (TR) described in Sec.\ref{sec:exp}.
All these tasks, when approached with NLI strategy, are tackled by comparing an input text (e.g., ``We really enjoyed the food, it is tasty and cheap, the staff was very nice and kind. However the restaurant is very hard to reach.'') and an hypothesis about the input text (e.g., ``The position of the restaurant is difficult to be reached'') and predicting if the input text either entails, contradicts, or is not related to the hypothesis (``Entailment'' is the correct prediction in the previous example). 
A common problem for many NLP tasks is the fact that the developed models usually require a big amount of natural language productions data, and usually they are made available in the English language. 
There are many languages that are underrepresented in these NLP dataset contexts and this made interesting tools hard to develop in these other languages, and there is the need to solve this to make these advance in tech available for low represented languages too. 
Data scarcity may be tacked with different strategies and this work describe some of them relatively to the NLI task in the Italian language. 
The goal of this research is to build a model with the following traits:
\begin{enumerate}[(a)]
\item 
it can perform the NLI task in a specific language;
\item based on the sentence embedding operation, such as in~\cite{SENTENCEBERT}; 
\item it is able to understand another language, in this case Italian; 
;
\item it is able of being general and not requiring any re-train for each specific industrial task (ABSA, SA, TR); 
\end{enumerate}
We can state the Research Question (RQ) which drive our effort is

\RQBOX{RQ: It is possible to build a NLI model with acceptable performances on NLI related downstream tasks in Italian language compliant with \textit{(a)}, \textit{(b)}, \textit{(c)}, \textit{(d)} constraints, without requiring a language-specific dataset?}

The main differences among the proposed models to achieve these aims is the training approach: one is based on Knowledge Distillation (KD)\cite{KD} a technique which aim to transfer knowledge from a \emph{Teacher} model (English-based NLI model) to \emph{Student} model (that will handle the Italian language).
The other model includes a step for the dataset translation from English to Italian by means of a Machine Translation model \cite{NLLB}. 
The approach exploiting KD has been demonstrated to have NLI capability in the target language, namely Italian, without being exposed to a NLI training dataset in Italian. This model has been named \textbf{I-SPIn} (\textbf{I}talian-\textbf{S}entence \textbf{P}air \textbf{In}ference) and is available at this \href{https://huggingface.co/Lutech-AI/I-SPIn}{link} along with all instructions for usage. 

The remainder of this paper is structured as follows: in Sec.~\ref{RelatedWork} we report literature and datasets discussion, Sec.~\ref{Method} describes the two implemented approaches, Sec.~\ref{sec:exp} reports the settings and results of the performed experiments, Sec.~\ref{sec:discussion} reports discussion and conclusions about this work.
\section{Related Work} 
\label{RelatedWork}
Common approaches to tackle NLI include Neural Networks, such as Recurrent Neural Networks or Transformer based methods~\cite{BILSTM,BERT}.
\cite{BILSTM} presents an architecture based both on learning \emph{Hypothesis} and \emph{Premise} in a dependent way using bidirectional LSTM and Attention mechanism~\cite{attention-original} to extract the text pair representation needed for final classification. The obtained results on SNLI~\cite{SNLI} validation set gives a $89\%$ accuracy.
\cite{BERT} describes  language representation model (BERT) building a model based on Transformers~\cite{ATTENTION,TRANSFORMERS} and pre-training it in a bidirectional way, this has the aim to serve as a pre-trained model that can be fine-tuned on several different tasks including NLI. BERT is fine-tuned and tested on MNLI dataset~\cite{MNLI} and reaches around $86\%$ accuracy.
Recent researches~\cite{T5,XLNET} demonstrate that Transformers models~\cite{ATTENTION} are more suited for the NLI task, consistently surpassing neural models~\cite{TRANSFORMERS}. 
All of these high performance approaches mainly hold for English language as it is the language in which there is the higher data availability. 
Some multi-language NLI approaches are proposed in~\cite{hypernymy}, where cross-lingual training, multilingual training, and meta learning are attempted using a dataset extracted from Open Multilingual WordNet. The best model resulted to be the one exploiting meta learning and reached $76\%$ accuracy on True/False classification task of text pairs for the Italian language.
\cite{magnini-etal-2014-excitement} represents another work on multi-language NLI where the Excitement Open Platform is presented as open source software for experimenting in NLI related tasks. It has many linguistics and entailment components based on transformations between \emph{Premise} (P) and \emph{Hypothesis} (H), edit distance algorithms, and a classification using features extracted from P and H. Italian Language is tested on a manually translated RTE-3 dataset~\cite{giampiccolo-etal-2007-third} and the best model has $63\%$ accuracy. 
In the context of an Italian Textual Entailment competition the task Recognizing Textual Entailment (RTE) is proposed. It is similar to NLI task but it only contains two Entailment Yes/No classes. The competition's winner model is described in~\cite{Bos2009TextualEA} and it is based on a open source software EDITS based on edit distance reaching $71\%$ accuracy on the convention EVALITA 2009 dataset which is extracted from Wikipedia.
\cite{Pakray2012RecognizingTE} presents a model based on translation. The input texts can be in any language and are translated into English using a standalone machine translation system. The authors show that machine translation can be used to successfully perform the NLI related tasks or when P and H are provided in different languages. For Italian, it uses Bing translation and it is tested on EVALITA 2009 dataset reaching $66\%$ accuracy.
\subsubsection{Datasets}
\label{datasets}
The datasets used in this work are described in this paragraph and examples can be found in Appendix \ref{appendix:data}.

The Stanford NLI (SNLI)~\cite{SNLI} corpus is a collection of 570k human-written English sentence pairs manually labeled for balanced classification with the labels ``entailment'', ``contradiction'', and ``neutral'', supporting the task of NLI. 
The SNLI dataset presents the canonical dataset split --- consisting of train, validation, and test sets.

Multi-Genre NLI (MNLI)~\cite{MNLI} corpus is a crowd-sourced collection of 433k sentence pairs annotated with textual entailment information. The corpus is modeled on the SNLI corpus, but differs in that covers a range of \emph{genres} of spoken and written text. 
The train set is composed of sentences with the same genres: ``Telephone'', ``Fiction'', ``Government'', ``Slate'' and ``Travel''. 
The MNLI dataset supports a distinctive cross-genre generalisation evaluation. There is a matched validation set which is derived from the same source as those in the training set, and a mismatched validation set which do not closely resemble any genres seen at training time.

RTE datasets (RTE3-ITA and RTE2009)  are English-native NLI datasets, manually translated by a community of researchers. 
The Italian version RTE3-ITA refers to the third refinement of this dataset
\footnote{The validation and test datasets can be downloaded at this \href{https://github.com/gilnoh/RTEFormatWork/tree/master}{link}.}.
Instead, RTE2009 was submitted for the EVALITA 2009 Italian campaign \cite{EVALITA}\footnote{The validation and test datasets can be found at this \href{https://www.evalita.it/campaigns/evalita-2009/data-distribution/}{link}.}.
These datasets are only used for testing, since they contain too few observations to be suitable for training. 
The RTE3-ITA dataset contains 1600 observations, whereas RTE-2009 contains 800 observations.
Unlike classical NLI, these datasets present only two labels: ``Entailment'' and ``No-Entailment''.

TED2020~\cite{TED2020} is a generic translation dataset. The option (English--Italian) has been selected for training among more than a hundred possible languages. The dataset consists of more than 400k parallel sentences. The transcripts have been translated by a community of volunteers. This dataset is used to make a model understand different languages~\cite{KD}, starting from a language known to the model. 

\section{Method}
\label{Method}
Three different architectures will be detailed throughout the section.
In Sec.~\ref{NLI training in the source language} the objective is to obtain the a model that is able to perform NLI in English. Starting from this model, we propose two parallel approaches to perform NLI in the target language. One is detailed in Sec.~\ref{Knowledge Distillation in the target language}, and the other is detailed in Sec.~\ref{Machine Translation in the target language}. Both approaches attempt a domain adaptation and generalisation in the target language --- namely, Italian, while lacking a language-specific dataset.
The models' parameters were selected among few different possibilities suggested by online informal documentation and literature. No cross-validation or grid-search analyses have been performed for computational constraints. Therefore, no guarantees on the optimality of the parameters can be made.
To reduce computational complexity during the inference phase for the models described in Sec.~\ref{Knowledge Distillation in the target language} and Sec.~\ref{Machine Translation in the target language}, we recommend to split the model to obtain independent instances of encoder and classifier. The proposed methodology is the following: transform all the sentence pairs in vectorial forms --- with the encoder --- first; in a second phase, the classifier will receive the embeddings to return an inference relation.
\subsubsection{NLI training in the source language} \label{NLI training in the source language}
The proposed solution makes use of a Transformer~\cite{ATTENTION}. The Transformer lately has become the state-of-the-art architecture for NLP, as detailed in \ref{RelatedWork}.
The first step of our methodology is to retrieve a sentence encoder model, based on Transformers. This sentence encoder model is already fine-tuned for general purposes over different languages. 
The encoder of choice to transform sentences in vectors was Sentence-BERT~\cite{SENTENCEBERT}. It is a fine-tuning of BERT~\cite{BERT}, that is a word embedding Transformer model, tailored for the task of sentence embedding. 
It has the ability to perform sentence embedding faster than BERT as detailed in~\cite{SENTENCEBERT}\footnote{Sentence-BERT was downloaded from this \href{https://huggingface.co/sentence-transformers/paraphrase-multilingual-mpnet-base-v2}{link}.}, by means of a Siamese training approach~\cite{SIAMESE}. 
Referring to this model with the term Sentence-BERT is inappropriate, since it has been fine-tuned on RoBERTa~\cite{ROBERTA}, that is a larger counterpart of BERT. Hence, the name Sentence-RoBERTa would be more appropriate. 
In this paper we will adopt the name Sentence-BERT to refer to any siamese structure accepting a sentence pair as input, including the instance of Sentence-RoBERTa to be fine-tuned.
Since Transformers are computationally expensive to train from scratch, we decided to test a multilingual version of Sentence-BERT and fine-tune it on SNLI and MNLI merged together to create a single NLI dataset.
After a fine-tuning session over the merged NLI dataset, the result is a model based on Transformers, that can proficiently address the NLI task --- only in English though, despite being originally trained on multiple languages. More information about this work available at~\cite{MULTILINGUAL}.

The output of the fine-tuned Sentence-BERT is composed of an embeddings pair, containing a vectorial representation of the premise and the hypothesis. Note that the sentence encoder model is invoked two separate times for this operation, for complexity optimisation reasons.
The Sentence-BERT output embeddings have been further transformed to maximise and emphasise the relevant information for our task. In detail, the following operations have been applied:
\begin{itemize}
    \item Element-wise product. Captures similarity of the two embeddings, and highlights components of the embeddings that are more relevant than others.
    \item Difference. Asymmetric operation; captures the direction of implication. We want the hypothesis to imply the premise, and not vice-versa.
\end{itemize}

\begin{figure}
  \centering
\includegraphics[width=0.75\textwidth, keepaspectratio]{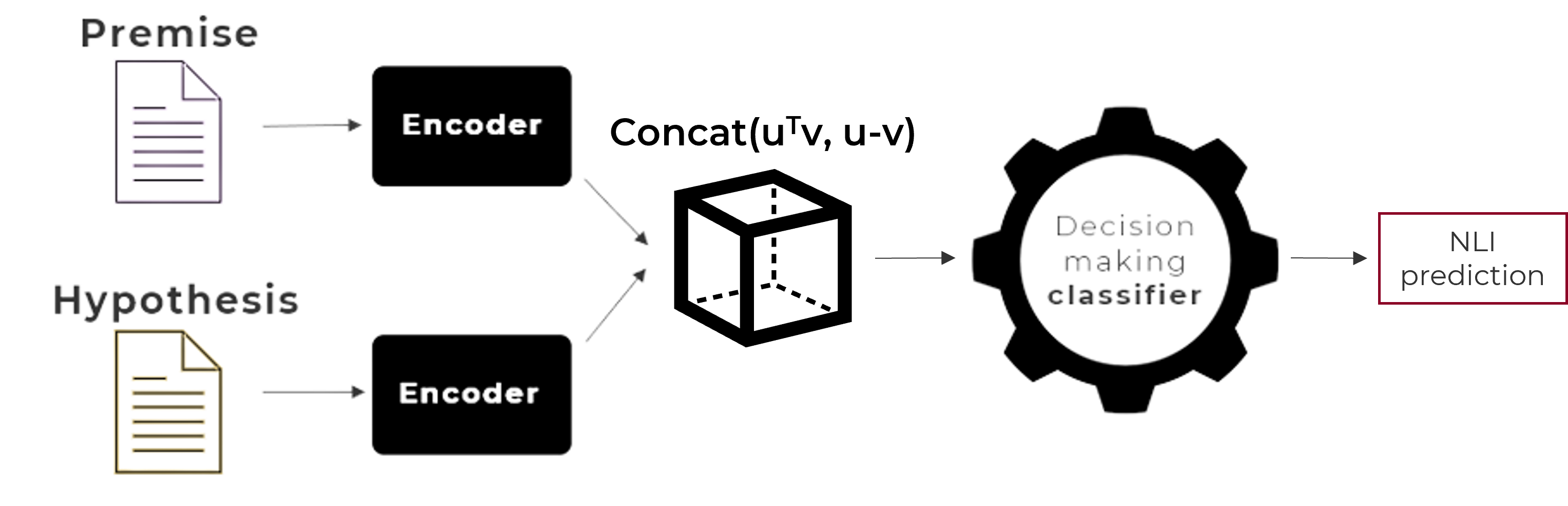}
  \caption{\label{Figure 1} Model structure. Two sentences are transformed in embeddings. The embeddings are compared with a classifier to get the prediction for the sentence pair.}
\end{figure}

The two transformed embeddings were concatenated and passed as input to a fully-connected Feed Forward architecture of six (6) layers, detailed in Appendix \ref{appendix:param}, with three (3) outputs (``Entailment'', ``Neutral'', ``Contradiction''), to predict the probability of the sentence pair to belong to each NLI class.
Finally, a softmax function was applied to the three-dimensional vector to obtain the class probabilities (Fig.~\ref{Figure 1}).

Execution-wise, the NLI fine-tuning task on a Tesla P100--PCIE--16GB GPU was completed in approximately six (6) hours on the merged NLI training dataset composed of an ensemble of SNLI and MNLI datasets, accounting for more than 1M observations. The main parameters can be found in Appendix \ref{appendix:param}.
In our work, we want to enable a multilingual Transformers-based model, previously fine-tuned for a specific task only in one specific language, to proficiently address that specific task in another language.

\subsubsection{Knowledge Distillation in the target language} \label{Knowledge Distillation in the target language}
The second step of our methodology is to employ a training without language-specific NLI training data and we selected the Knowledge Distillation (KD)~\cite{KD} approach.

KD was born as a model compression technique~\cite{HINTON}, where knowledge is transferred from the teacher model to the student by minimizing a loss function, in which the target is the distribution of class probabilities predicted by the teacher model. 
KD is a powerful technique since it can be used for a variety of multiple tasks. In our experiments, we employed KD to perform NLI in the target language, with the objective of forcing a translated sentence to have the same embedding --- i.e. location in the vector space --- as the original sentence. 
The soft targets of the teacher model constitute the labels to be compared with the predictions returned by the student model. 
The task at hand may fall in the domain adaptation problem sphere.

We require a teacher model (encoder) $T$, that maps sentences in the source language to a vectorial representation. 
Further, we need parallel (translated) sentences $D = ((source_1, target_1), ...,(source_n, target_n))$ with $source_j$ being a sentence in the source language and $target_j$ being a sentence in the target language. We train a student encoder model $S$ such that $T(source_j) \approx S(target_j)$. 
For a given mini-batch $B$, we minimise the Mean Squared Error loss function:

\begin{equation}
MSE_{(S, T, D = (source_j, target_j))} = \frac{1}{|B|}\sum_{j \in |B|} ( T(source_j) - S(target_j) )^2 \\
\end{equation}

\begin{figure}
  \centering
\includegraphics[width = 0.75\linewidth]{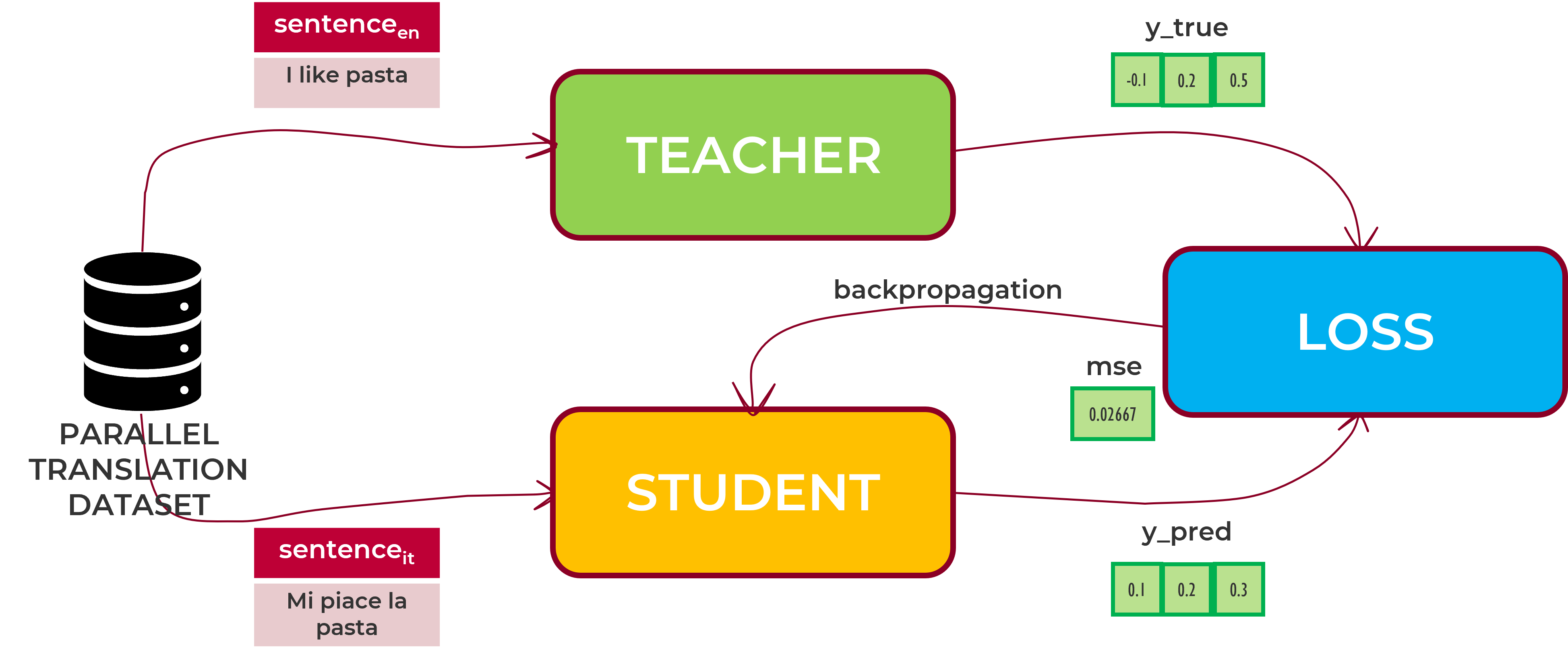}
  \caption{\label{Figure 2} Knowledge Distillation. Teacher encoder model receives source sentences, student model receives target sentences. Student encoder model is updated with new information from the teacher.}
\end{figure}

Two instances of the encoder described in Sec.~\ref{NLI training in the source language} have been taken for the experiment. One acts as teacher encoder model $T$, the other as a student encoder model $S$.
The application of KD has the objective to share the domain knowledge of the teacher encoder model to the student encoder model, and at the same time learn a new vectorial representation for the target language. 
A schematic representation is provided in Fig.~\ref{Figure 2}. 

The obtained NLI classifier, able to understand Italian, accepts a sentence pair to output a NLI label.
Execution-wise, the KD task on a Tesla P100--PCIE--16GB GPU was completed in approximately five (5) hours on the TED2020 (English--Italian) dataset consisting of more than 400k parallel sentences. The main parameters can be found in Appendix \ref{appendix:param}.
\subsubsection {Machine Translation in the target language} \label{Machine Translation in the target language}
As an alternative method for our second step, we employ a Large Language Model named No Language Left Behind (NLLB)~\cite{NLLB} to address the lack of language-specific NLI training data. To the best of our knowledge, it was not possible to find a comprehensive NLI dataset in Italian. The RTE3-ITA and RTE-2009 datasets, both detailed in Sec.~\ref{datasets}, together present about 2500 observations, too few to train a Deep Learning model. Therefore, the dataset used to fine-tune this architecture is the same as in  Sec.~\ref{NLI training in the source language}, with an alteration: we perform a translation of the dataset. In fact, the simplest way to perform NLI in a language other than English is to machine translate the ensemble NLI dataset, consisting in SNLI and MNLI merged together.
Note that, for memory and performance optimisation, the ensemble NLI training dataset was dynamically translated during execution by invoking the NLLB model for each mini-batch.
Execution-wise, this fine-tuning task over the target language on a Tesla P100--PCIE--16GB GPU was completed in approximately ten (10) hours on the translated ensemble NLI dataset, consisting of more than 1M sentence-pairs.  The main parameters can be found in Appendix \ref{appendix:param}.

\section{Experiments}
\label{sec:exp}
\subsubsection{NLI results in the source language} \label{NLI results in the source language}
The architecture discussed as in  Sec.~\ref{NLI training in the source language} has been tested over the standard NLI task in English.
For SNLI the accuracy reached $80.69\%$ while for MNLI a $77.00\%$ accuracy is reached. 
\subsubsection{NLI results in the target language} \label{NLI results in the target language}
The architecture discussed as in Sec.~\ref{Knowledge Distillation in the target language} --- that is the main focus of this paper --- has been tested over the standard NLI task in Italian, and compared with the alternative architecture based on Machine Translation. The underlying model, an open-source machine translation model, developed by Facebook, named No Language Left Behind~\cite{NLLB}, was also exploited to obtain a comprehensive Italian NLI dataset, suitable for testing.
Results for the SNLI and the MNLI test sets (both translated in Italian) are detailed in Tab.~\ref{nli_ita_res}.
\begin{table}
\centering
\label{nli_ita_res}
 \caption{ NLI (IT) results.}
\begin{tabular}{|c|c|c|c|c|}
\hline
Dataset  &Task& Acc. & Min F1 & Macro-Avg F1 \\ \hline
SNLI (IT) & NLI & 74.21 (-1.83\% ) &  67.19\% (-4.34\% ) & 74.08\% (-4.94\%) \\ \hline
MNLI-Mismatch (IT) &
   NLI &
72.74\% (\textbf{+1.09\%})&
 64.53\% (\textbf{+0.55\%})
 & 72.78\% (\textbf{+1.37\%}) \\ \hline

 RTE3-ITA & RTE & 67.50\% (\textbf{+4.75\%})
  & 60.12\% (\textbf{+5.55\%}) & 66.35\% (\textbf{+4.85\%}) \\ \hline
RTE-2009 & RTE & 59.00\% (-0.75\%) 
& 31.09\% (-2.65\%) 
& 50.96\% (-1.46\%)\\ \hline
\end{tabular}%
\end{table}

SNLI results in Tab.~\ref{nli_ita_res} are not far from the theoretical accuracy cap these models have --- presented for NLI in source language. 
This could be interpreted as a success for the training of both architectures.
The Min F1-Score metric captures the most misclassified class. The Neutral class, in general, has been the most challenging to classify, as translation biases may slightly change the connotation of a sentence.
Note that this test is biased towards the Machine Translation-based architecture. Remember that this architecture has been fine-tuned over the translated NLI dataset in the target language; 
the KD-based architecture, instead, had never seen the NLI dataset in the target language. This suggests that the KD-based architecture may have relevant learning capabilities over this task.
Differently from the SNLI dataset, we briefly remark that MNLI datasets are divided into genres, and supports a distinctive cross-genre generalisation evaluation by means of the mismatched validation set. A higher accuracy on the mismatched validation set corresponds to a better generalisation of the model.
In the same way as before, also for this test the Machine Translation-based architecture had an objective advantage, by being trained on the same dataset it was tested over. Nonetheless, the KD-based architecture performed better in this test. This dataset tests the generalisation capability and the ability to understand a wide range of contexts of a model, as it contains multiple genres. This could be a motivation to consider the KD-based architecture the most powerful architecture of the two.

In addition to the tests above, the architecture has been tested over the RTE datasets.
We briefly remind that, unlike classical NLI, these datasets present only two labels: ``Entailment'' and ``No-Entailment''. Both our models produce three labels --- ``Entailment'', ``Neutral'', ``Contradiction'' --- as they were trained on SNLI and MNLI. 
The two-label mapping for this task maps both \emph{Neutral} and \emph{Contradiction} to \emph{No-Entailment}
 as this maximises the accuracy on the validation set.
Results for the RTE3-ITA and RTE-2009 test sets are reported in Tab.\ref{nli_ita_res} too.
The performance difference between the two architectures may be explained by the difference in quality of the target language the two architectures have been exposed to during training. In fact, Machine Translation-based architecture has been trained over a machine translated dataset, whereas the KD-based architecture was trained over a manually translated dataset. This supposition can be made because this dataset has been manually translated in Italian, hence presents a better language quality than the NLI datasets translated in the target language.

\subsubsection{ABSA results} \label{ABSA results}
Aspect-Based Sentiment Analysis at EVALITA (ABSITA), detailed in~\cite{EVALITA}, is an ABSA dataset. 
Contains Italian hotel reviews that may touch different topics (such as price, location, cleanliness, etc.) and a sentiment associated to each topic (knowing that sentiments for different topics may be contrasting). 
By choosing arbitrary NLI hypotheses, this dataset may emulate a total of three (3) different tasks, namely SA, TR, and ABSA. 
The core idea behind this setting comes from the desire to query a text --- in NLI, a set of premises (e.g. a set of reviews), in an unsupervised way, to receive specific answers from a predefined list of answers (e.g. the presence of a topic from a list of topics). 
In the case of open answers, a question-answer architecture would have been more suitable.
\begin{table}
\centering
\caption{\label{absita res} ABSITA results, over the Sentiment Analysis and Topic Recognition tasks}
\begin{tabular}{|c|c|c|c|c|c|}
\hline
Dataset & Balancing &Task& Acc. & Min F1 & Macro-Avg F1 \\ \hline
ABSITA  & 1:1 & SA&  88.12\% (\textbf{+3.08\%} )  & 86.89\% (\textbf{+3.19\%})    &  
88.02\% (\textbf{+3.09\%}) \\ \hline
ABSITA  & 1:1 & TR&  68.09\% (-3.1\% )  & 65.75\% (-3.01\%)    &  
67.97\%  (-3.16\% ) \\ \hline
ABSITA  & 1:7 & TR&  71.11\%  (\textbf{+5.27\%} )  & 37.94\% (-0.77\%)    &  
59.56\%   (\textbf{+2.04\%} ) \\ \hline
ABSITA & 1:1  & ABSA &  94.03\% (\textbf{+6.24\%}) &  93.90\% (\textbf{+6.65\%})& 94.02\% (\textbf{+6.35\%}) \\ \hline
ABSITA & 1:15 & ABSA & 78.42\% (\textbf{+11.39\%})  & 37.66\% (\textbf{+8.3\%}) & 62.30\% (\textbf{+8.37\%}) \\ \hline
\end{tabular}%
\end{table}

\textit{Sentiment Analysis}
 (SA) is the task to recognise the overall sentiment of a sentence. As detailed above, we would like to exploit the models to apply SA in an unsupervised manner --- to do this, we fix a hypothesis arbitrarily. We assume that the hypothesis we have chosen captures the logical implication that is the core of NLI.
 Results for the ABSITA dataset are detailed in Tab.~\ref{absita res}. Note that the hypothesis has been arbitrarily set to ``Sono soddisfatto'' (``I feel satisfied''), hence ``Entailment'' refers to the model predicting positive sentiment.
The two-label mapping for this task maps \emph{Neutral} to \emph{Entailment}.

\textit{Topic Recognition}
 (TR) is the task to recognise whether or not a sentence is about a topic. As detailed above, we would like to exploit the models to apply TR in an unsupervised manner --- to do this, we fix a hypothesis arbitrarily. We assume that the hypothesis we have chosen captures the logical implication that is the core of NLI.
 Results for the ABSITA dataset are detailed in Tab.~\ref{absita res}. 
The seven (7) in the ``Balancing'' column stands for the number of different topics in the dataset. The 1:1 balancing has been obtained by randomly sampling sentences from the seven (7) classes that do not compose the target. The two scenarios have been proposed to extensively test the generalisation capability of the models.
Note that the hypothesis has been arbitrarily set to ``Parlo di pulizia'' (``I'm talking about cleanliness''), hence ``Entailment'' refers to the model predicting the label ``cleanliness''.
The two-label mapping for this task maps \emph{Neutral} to \emph{Entailment}.
\textit{Aspect-Based Sentiment Analysis}
 (ABSA) is the task to recognise the sentiment about each sub-topic in a sentence. As detailed above, we would like to exploit the models to apply ABSA in an unsupervised manner --- to do this, we fix a hypothesis arbitrarily. We assume that the hypothesis we have chosen captures the logical implication that is the core of NLI.
Results for the ABSITA dataset are detailed in Tab.~\ref{absita res}. 
Note that the hypothesis has been arbitrarily set to ``La camera \'e pulita'' (``The room is clean''), hence ``Entailment'' refers to the model predicting positive sentiment and ``cleanliness'' label.
The two-label mapping for this task maps \emph{Neutral} to \emph{Contradiction}.

\section{Conclusions and Discussions}
\label{sec:discussion}
To interpret the apparently decent results for NLI in the source language, listed in  Sec.~\ref{NLI results in the source language}, we need to consider the fact that, during training, sentence encoders do not look at both inputs simultanously, hence generating good but not top-tier performances. 
Potentially, we could have obtained slightly better results by making use of a word encoder instead of a sentence encoder, at the cost of a large computational overhead.
To address various industrial tasks, we decided to prioritise scalability and responsiveness.
The discussed architecture, based on KD, demonstrated to perform better than the other architecture --- that was directly trained over machine translated NLI datasets --- despite having an objective disadvantage. 
We stress the fact that the proposed architecture was never directly trained over any kind of Italian NLI data.
Compared to the other methodology, the KD presents the following advantages: 
\begin{enumerate}
    \item Easier to extend models: we just require few samples for the new languages.
    \item Lower hardware requirements: machine translation --- that is an expensive task --- is not needed as an intermediate step.
\end{enumerate}

To test our model performances over SA, TR, and ABSA, we employed arbitrary hypotheses. 
We tried our best to avoid any biases (e.g. hypotheses were chosen by colleagues that had never taken a look at the datasets), but we acknowledge that some bias may have been introduced. This is currently considered an open problem.

Different architectures have been tested showing that it is possible to obtain reasonable accuracies over different NLP tasks by fine-tuning a single architecture based on sentence embeddings over the NLI task. 
We showed that various NLP problems may be mapped into a NLI task --- in this way, we empirically proved the generality of the NLI task. 
We would like to stress over the lack of need to re-train any models to obtain the results over each specific task.
Moreover, lately NLI models find an important academic usage for boosting the consistency and accuracy of NLP
models without fine-tuning or re-training~\cite{CONCORD}. This is because models should demonstrate internal self-consistency, in the sense that their predictions across inputs should imply logically compatible beliefs about the world --- NLI models are trained to achieve that understanding.
\bibliographystyle{splncs04}
\bibliography{asdfromeyetracks}
\newpage
\appendix
\section{Dataset examples}
\label{appendix:data}
Examples from the benchmark Stanford NLI dataset are shown in Tab.~\ref{snli table} to show its standard structure.
\begin{table}%
[H]
  \centering
  \label{snli table} 
  \caption{Stanford NLI dataset. A label is produced based on the logical interaction between two short texts.}
  \begin{tabular}{|c|c|c|}
    \hline
    Premise & Hypothesis & Label \\ \hline
    \begin{tabular}[c]{@{}c@{}}``A soccer game with \\ multiple males playing''\end{tabular} &
    \begin{tabular}[c]{@{}c@{}}``Some men are \\ playing a sport''\end{tabular} & Entailment \\ \hline
    \begin{tabular}[c]{@{}c@{}}``An older and younger \\ man smiling''\end{tabular} &
    \begin{tabular}[c]{@{}c@{}}``Two men are smiling\\  and laughing at the cats\\  playing on the floor''\end{tabular} & Neutral \\ \hline
    \begin{tabular}[c]{@{}c@{}}``A man inspects the\\  uniform of a figure \\ in some East Asian country''\end{tabular} & ``The man is sleeping'' & Contradiction \\ \hline
  \end{tabular}%
\end{table}

Examples from the benchmark Multi-Genre NLI dataset are shown in Tab.~\ref{mnli table} to show its standard structure.
\begin{table}
  \centering
  \label{mnli table} 
  \caption{Multi-Genre NLI dataset. A label is produced based on the logical interaction between two short texts.}
  \begin{tabular}{|c|c|c|}
    \hline
    Premise & Hypothesis & Label \\ \hline
    \begin{tabular}[c]{@{}c@{}}``It has a staff of about 100 employees, \\ including attorneys and support staff, \\ in 10 branch offices.''\end{tabular} &
    \begin{tabular}[c]{@{}c@{}}``The 10 branches had close \\ to 100 employees.''\end{tabular} & Entailment \\ \hline
    \begin{tabular}[c]{@{}c@{}}``Theoretically scale economies in \\ delivery are not firm specific.''\end{tabular} &
    \begin{tabular}[c]{@{}c@{}}``Scale economies are flexible.''\end{tabular} & Neutral \\ \hline
    \begin{tabular}[c]{@{}c@{}}``Mrs. Cavendish is in her \\ mother-in-law's room. ''\end{tabular} &
    \begin{tabular}[c]{@{}c@{}}``Mrs. Cavendish has left \\ the building.''\end{tabular} & Contradiction \\ \hline
  \end{tabular}%
\end{table}

Examples from the RTE3-ITA NLI dataset are shown in Tab.~\ref{rte tab}. Note the table will be proposed in the Italian language.
\begin{table}
  \centering
  \label{rte tab} 
  \caption{RTE3-ITA dataset. A label is produced based on the logical interaction between two short texts.}
  \begin{tabular}{|c|c|c|}
    \hline
    Premise & Hypothesis & Label \\ \hline
    \begin{tabular}[c]{@{}c@{}}``All'uscita del gioco \\ Final Fantasy III nella versione \\ per la console Super Nintendo, \\ il nome di Bigg era Vicks.''\end{tabular} &
    \begin{tabular}[c]{@{}c@{}}``Final Fantasy III venne \\ prodotto per la console \\ Super Nintendo.''\end{tabular} & Entailment \\ \hline
    \begin{tabular}[c]{@{}c@{}}``La signora Minton lasciò \\ l'Australia nel 1961 per \\ proseguire i suoi studi a Londra.''\end{tabular} &
    \begin{tabular}[c]{@{}c@{}}``La signora Minton è \\ nata in Australia.''\end{tabular} & Contradiction \\ \hline
  \end{tabular}%
\end{table}

Examples from the TED2020 translation dataset are shown in Tab.~\ref{ted}.
\begin{table}
\centering
\resizebox{\columnwidth}{!}{%
\begin{tabular}{|c|c|}
\hline
sentence$_{en}$ & sentence$_{it}$ \\ \hline
\begin{tabular}[c]{@{}c@{}}``I gave my speech, then went back \\ to the airport to fly back home.''\end{tabular} &
  \begin{tabular}[c]{@{}c@{}}``Io feci il mio discorso, poi \\ andai all'aeroporto per tornare.''\end{tabular} \\ \hline
\begin{tabular}[c]{@{}c@{}}``He romanticised the idea \\ they were star-crossed lovers.''\end{tabular} &
  \begin{tabular}[c]{@{}c@{}}``Lui fantasticava sull'idea di \\ loro come amanti sfortunati.''\end{tabular} \\ \hline
\begin{tabular}[c]{@{}c@{}}``In Japan, a game of ping-pong \\ is really like an act of love.''\end{tabular} &
  \begin{tabular}[c]{@{}c@{}}``In Giappone, una partita di ping-pong\\  \'e come un atto d'amore.''\end{tabular} \\ \hline
\end{tabular}%
}
\caption{\label{ted} TED2020 dataset (English--Italian version).}
\end{table}

\section{Models parameters}
\label{appendix:param}
Fully-connected Feed Forward architecture used for classification in Sec.~\ref{NLI training in the source language} is reported here:

\noindent
\begin{Verbatim}[tabsize=4]
(layers): ModuleList(
     (0): Linear(in=1536, out=1024, activation=GELU())
     (1): Linear(in=1024, out=512, activation=GELU())
     (2): Linear(in=512, out=256, activation=GELU())
     (3): Linear(in=256, out=128, activation=GELU())
     (4): Linear(in=128, out=64, activation=GELU())
     (5): Linear(in=64, out=3, activation=GELU())
)
\end{Verbatim}

Hyper-parameters for NLI model training in the source language Sec.~\ref{NLI training in the source language} are listed here:

\begin{itemize}
\item \texttt{batch\_size = 8}
\item \texttt{max\_sentence\_length = 256}
\item \texttt{max\_tokens\_length = 128}
\item \texttt{epochs = 1}
\item \texttt{learning\_rate = 2e-5}
\item \texttt{epsilon = 1e-8}
\item \texttt{weight\_decay = 0}
\item \texttt{accumulation\_step = 8}
\end{itemize}

Hyper-parameters for NLI model training in the source language Sec.~\ref{Knowledge Distillation in the target language} are listed here:

\begin{itemize}
\item \texttt{batch\_size = 24}
\item \texttt{max\_sentence\_length = 256}
\item \texttt{max\_tokens\_length = 128}
\item \texttt{epochs = 6}
\item \texttt{learning\_rate = 2e-5}
\item \texttt{epsilon = 1e-6}
\item \texttt{weight\_decay = 1e-2}
\item \texttt{accumulation\_step = 4}
\end{itemize}

Hyper-parameters for NLI model training in the source language Sec.~\ref{Machine Translation in the target language} are listed here:

\begin{itemize}
\item \texttt{batch\_size = 8}
\item \texttt{max\_sentence\_length = 256}
\item \texttt{max\_tokens\_length = 256}
\item \texttt{epochs = 5}
\item \texttt{learning\_rate = 4e-5}
\item \texttt{epsilon = 1e-16}
\item \texttt{weight\_decay = 1e-4}
\item \texttt{accumulation\_step = 4}
\end{itemize}

\end{document}